\documentclass[sigconf]{acmart}
\pdfoutput=1

\AtBeginDocument{%
  \providecommand\BibTeX{{%
    \normalfont B\kern-0.5em{\scshape i\kern-0.25em b}\kern-0.8em\TeX}}}

\usepackage{colortbl}
\usepackage{multicol}
\usepackage{multirow}
\usepackage{makecell}
\usepackage{diagbox}
\usepackage{booktabs}
\usepackage{xcolor}
\usepackage{subfig}
\usepackage{graphicx}
\usepackage{subfloat}
\newcommand\crule[3][black]{\textcolor{#1}{\rule{#2}{#3}}}
\usepackage{enumitem}
\setlist{nosep}
\usepackage{balance}

\setlist[itemize]{leftmargin=*}
\AtBeginDocument{%
  \providecommand\BibTeX{{%
    \normalfont B\kern-0.5em{\scshape i\kern-0.25em b}\kern-0.8em\TeX}}}

\copyrightyear{2021}
\acmYear{2021}
\setcopyright{acmcopyright}\acmConference[SIGIR '21]{Proceedings of the 44th International ACM SIGIR Conference on Research and Development in Information Retrieval}{July 11--15, 2021}{Virtual Event, Canada}
\acmBooktitle{Proceedings of the 44th International ACM SIGIR Conference on Research and Development in Information Retrieval (SIGIR '21), July 11--15, 2021, Virtual Event, Canada}
\acmPrice{15.00}
\acmDOI{10.1145/3404835.3462945}
\acmISBN{978-1-4503-8037-9/21/07}

\begin{document}
\fancyhead{}
\pagenumbering{gobble}
\title{Legal Judgment Prediction with Multi-Stage Case Representation Learning in the Real Court Setting
}

\author{Luyao Ma$^{1,2}$, Yating Zhang$^{2}$, Tianyi Wang$^{2}$, Xiaozhong Liu$^{3\dagger}$, Wei Ye$^{1\dagger}$}
\author{Changlong Sun$^{2}$, Shikun Zhang$^{1}$}
\affiliation{
  \institution{$^1$National Engineering Research Center for Software Engineering, Peking University}
}
\affiliation{
  \institution{$^2$Alibaba Group}
}
\affiliation{
  \institution{$^3$Indiana University Bloomington}
}
\email{1701210338@pku.edu.cn;ranran.zyt@alibaba-inc.com;will.wty@alibaba-inc.com}
\email{liu237@indiana.edu;wye@pku.edu.cn;changlong.scl@taobao.com;zhangsk@pku.edu.cn}

\begin{abstract}
Legal judgment prediction(LJP) is an essential task for legal AI. While prior methods studied on this topic in a pseudo setting by employing the judge-summarized case narrative as the input to predict the judgment, neglecting critical case life-cycle information in real court setting could threaten the case logic representation quality and prediction correctness. In this paper, we introduce a novel challenging dataset\footnote{https://github.com/mly-nlp/LJP-MSJudge} from real courtrooms to predict the legal judgment in a reasonably encyclopedic manner by leveraging the genuine input of the case -- plaintiff's claims and court debate data, from which the case's facts are automatically recognized by comprehensively understanding the multi-role dialogues of the court debate, and then learnt to discriminate the claims so as to reach the final judgment through multi-task learning. An extensive set of experiments with a large civil trial data set shows that the proposed model can more accurately characterize the interactions among claims, fact and debate for legal judgment prediction, achieving significant improvements over strong state-of-the-art baselines. Moreover, the user study conducted with real judges and law school students shows the neural predictions can also be interpretable and easily observed, and thus enhancing the trial efficiency and judgment quality. 
\end{abstract}


\keywords{judgment prediction, case life-cycle, multi-task learning\renewcommand{\thefootnote}{\fnsymbol{footnote}}\footnotetext[2]{Corresponding authors}}

\renewcommand{\thefootnote}{\arabic{footnote}}

\maketitle
\section{Introduction}
\begin{figure*}[!t]
\centering
\includegraphics[width=.8\linewidth]{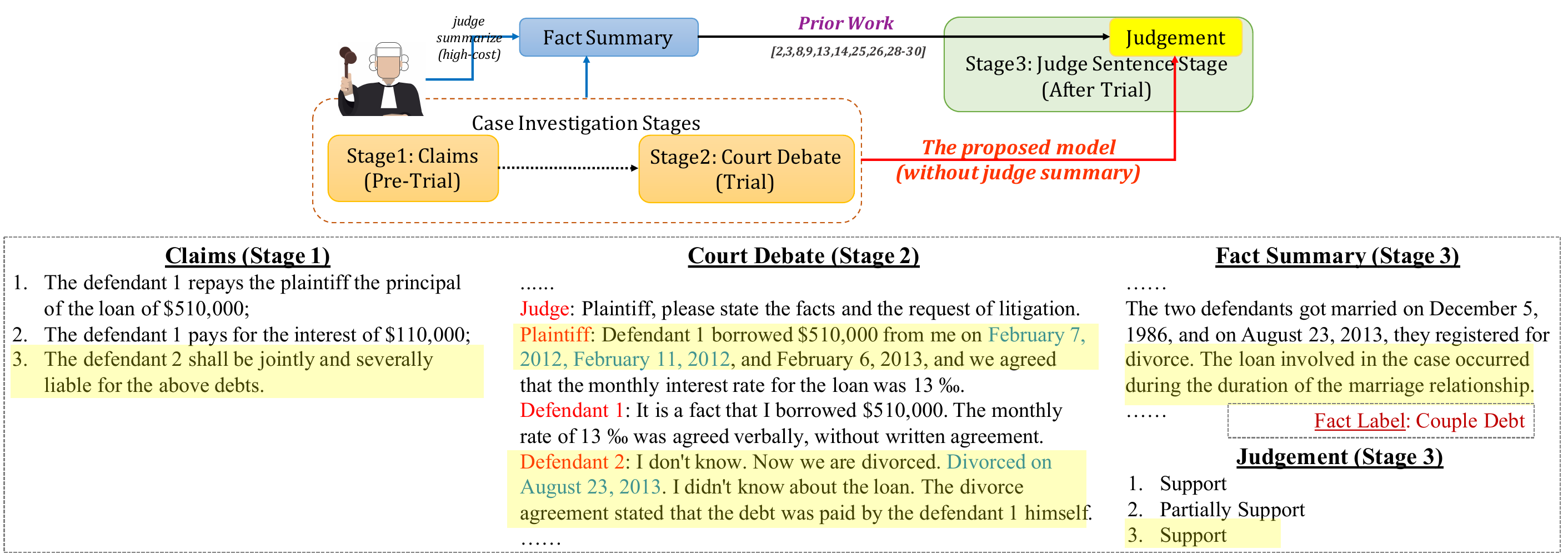}
\vspace{-4pt}
\caption{Conceptual overview of the difference between previous work and MSJudge.}
\vspace{-8pt}
\label{fig:approach_graph}
\end{figure*}

Argus, the Greek mythological giant with hundred eyes, could be an appropriate metaphor to characterize an ideal legal judgment predictor, which is expected to comprehensively examine the case from different views and across various stages of the case life-cycle. Neglecting any detailed information, e.g., admissibility of evidence, narrative from witnesses, and response from plaintiff/defendant, could mislead the prediction outcome. While legal judgment prediction has been originally proposed in 1960s (entitled ``Using Simple Calculations Predict Judicial Decisions \cite{nagel1960using}’’), unfortunately, prior studies \cite{chalkidis-etal-2019-neural,ye2018interpretable,jiang2018interpretable,zhong2018legal,luo2017learning,hu2018few,zhong2020iteratively,xu2020distinguish,yang2019legal,chen2019charge,long2019automatic,haar1977computer,lawlor1963computers,wiener1962decision} ignored the multi-stage nature of the legal case, instead researching in a pseudo setting which regards the judge-summarized case narrative as input to predict the judgment results. The neglect of critical case life-cycle information in real court setting, however, could threaten the case logic representation quality and prediction correctness. 

In this study, in order to recover the case jigsaw puzzle, we introduce a novel challenging dataset collected from real courtrooms as well as propose an innovative neural model to integrate pre-trial claims and court debate. More importantly, the case's facts are automatically recognized by comprehensively understanding the colloquial and lengthy court debate, and then learning to discriminate each claim so as to reach the final judicial conclusion.

For a litigation process, a case life-cycle often experiences three critical stages (depicted in Fig. \ref{fig:approach_graph}): pre-trial claim collection stage (e.g., plaintiff provides narrative to judge for the target case), trial court debate stage (e.g., plaintiff, defendant, witness, lawyer and judge debate on the court focusing on the claims), and after trial judge sentence stage (judge generates verdict, often including case fact summary and judgement). In the first and second stages, the judge usually spends high cost in identify legal facts from court debate transcript and evidence materials so as to further make final judgement in stage 3. Thus a comprehensive case life-cycle representation learning can be nontrivial for legal prediction.

To achieve this goal, the biggest challenge lies in the difficulty of accurately representing the multi-role court debate, where different camps may not necessarily share the same vocabulary space, and classical NLP algorithms can hardly consume this variation. For instance, the judge can be more responsible for investigating the facts and reading the court rules while the other litigants answer the questions from the judge. Moreover, with opposite position, plaintiff and defendant's attitudes, sentiments and descriptions to the same topic can be quite different. The second barrier comes from the gap between the litigants' statements during the court debate and the facts to be recognized by the judge. There exist much non-judicial content and noisy logic in court debate. As the exemplar case shown in Fig. \ref{fig:approach_graph}, whether the third claim should be supported or not is highly rely on whether the loan is occurred during the defendants' marriage (as described in fact summary), thus the model should learn to identify such fact from the court debate even though much noisy information (e.g., \textit{divorce}, \textit{paid by the defendant 1 himself}) may distract such recognition. Only the statements convincing enough with thorough evidences to be supported can be regarded as identified facts for further making judgment prediction. Therefore, discoverability of the truth/facts from colloquial dialogue content is an essential factor in the task of legal judgment prediction. The last but not the least is the challenge of representing the relations/interactions among the debate, facts, claims and judgment. In civil cases, the judgment can be generalized as the answer to the claims while it is common to have multiple claims in one case and whether they are established or not is not relatively independent.

Motivated by such observations, in this paper, we propose a novel neural judgment prediction model by addressing multi-stage representation learning called \textsf{MSJudge} (\textbf{M}ulti-\textbf{S}tage \textbf{Judge}ment Predictor). It jointly learns the identification of the legal facts in the court debate and simultaneously predicts the judgment result of each claim. As aforementioned example, due to the content and logic gap existing between court debate and fact summary, an auxiliary supervision by the fact enables to filter the noisy and non-judicial content from stage 1 and 2 for further better judgment prediction. Fig. \ref{fig:model_description} depicts the architecture of \textsf{MSJudge}. In a joint learning process, various kinds of information collected from different stages, e.g., court debate, facts, claims, are encapsulated to regularize the judgment prediction of the civil case through the fine-tuning of the court debate representation. To make the prediction results interpretable and being easily observed, we provide visualization for accessing the mutual influence between different components. 

In summary, this paper makes the following key contributions: 
(1) We take a concrete step towards augmenting judicial decision making by investigating a novel task of judgment prediction through the case life-cycle data (Section \ref{Sec:preliminary}), which is essential for the practical use of AI techniques in a real court setting. 
(2) We propose an end-to-end framework called \textsf{MSJudge} (Section \ref{Sec:model}) that operates in a manner of multi-task supervision with multi-stage representation learning. The proposed model enables to address the judgment prediction by exploring and visualizing the interactions/mutual effect between ``debate and fact'', ``fact and claim'' and ``across claims''. It summons a practical scenario for legal judgement automation.
(3) \textsf{MSJudge} is trained in a supervised manner where the training data is extracted from over $70k$ court records of civil trials along with their judgment documents. Our experimental study (Section \ref{Sec:result}) demonstrates superiority of \textsf{MSJudge} over competitive baselines. The proposed method achieves $86.5\%$ in micro $F\_1$ score on the multi-stage trial datase, and significantly outperforms the best performing baselines (3.6\% increase in performance). Meanwhile, in the experiment, the proposed approach challenged upper bound idealization (pseudo setting), i.e., using judge summarized case fact as input. The equivalent prediction performance ($86.6\%$ in micro $F_1$ score) grants us confidence that the proposed model is able to optimize the case representation across different stages for more practical use in real court setting. Importantly, we also show its effectiveness in judicial 
decision making with a user study involving real law school students and judges (Section~\ref{Sec:user_study}).

\begin{table*}[!t]
\scriptsize
\caption{Comparison among several state-of-the-art work. Note that the meaning of the colors is the same as in Table 2. \crule[yellow]{0.3cm}{0.3cm}
\crule[green]{0.3cm}{0.3cm} \crule[blue]{0.3cm}{0.3cm} stand for employing the data at stage 1 (claim), stage 2 (court debate) and stage 3 (fact summary), respectively.\crule[gray]{0.3cm}{0.3cm} represents to employ the fact labels at stage 3 for supervision in training phase.}
\label{tab:related_work}
\begin{tabular}{|p{18mm}|p{30mm}|p{12mm}|p{80mm}|p{3mm}|p{3mm}|p{3mm}|}
\hline
\multicolumn{1}{|c}{Study}                                                                                                                              & \multicolumn{1}{|c}{Task}                                                                                                                  & \multicolumn{1}{|c}{Case Type}       & \multicolumn{1}{|c|}{Methodology}                                                                                                                                                                                                                        & \multicolumn{3}{c|}{Stage} \\ \hline
Few-Shot\cite{hu2018few}                                                                  & Charge Prediction                                                                                                   & Criminal Case &  an attribute-attentive
charge prediction model, which employs discriminative attributes to learn attribute-aware fact representation.                                                                        &         &         &     \cellcolor{blue}    \\
\midrule
TOPJUDGE\cite{zhong2018legal}                                                                              & \begin{tabular}[c]{@{}l@{}}Charge Prediction\\      Law Prediction\\      Term Prediction\end{tabular}              & Criminal Case & a topological multi-task learning framework for LJP, which formalize the
dependencies among subtasks as a Directed
Acyclic Graph.                                                                                    &         &         &  \cellcolor{blue}       \\
\midrule
LJP-QA\cite{zhong2020iteratively} & Charge Prediction                                                                                                   & Criminal Case & a reinforcement learning
method by iteratively questioning and answering to provide
interpretable results for legal judgment prediction.                                                                                                                    &         &         & \cellcolor{blue}        \\
\midrule
LADAN\cite{xu2020distinguish}                                                                 & \begin{tabular}[c]{@{}l@{}}Charge Prediction\\      Law Prediction\\      Term Prediction\end{tabular}              & Criminal Case & use a novel graph distillation operator (GDO) to extract discriminative features for effectively distinguishing confusing law articles.                                                                                               &         &         &  \cellcolor{blue}       \\
\midrule
Autojudge\cite{long2019automatic}                                                                    & Judgment Prediction                                                                                                 & Civil Case    & formalize the task as Legal Reading Comprehension according to the legal scenario.                                                                                                                                               & \cellcolor{yellow}        &         & \cellcolor{blue}        \\
\midrule
\textbf{MSJudge}                                                                                      & \begin{tabular}[c]{@{}l@{}}Judgment Prediction\\     Judicial Fact Recognition\end{tabular} & Civil Case  & release a case life-cycle dataset and propose a multi-task learning framework by leveraging multi-stage judicial data with consideration of the interactions among claims, court debate and fact recognition.                                                                                                          &  \cellcolor{yellow}        &   \cellcolor{green}       &   \cellcolor{gray}    \\\hline 
\end{tabular}
\vspace{-8pt}
\end{table*}
\vspace{-4pt}
\section{Related Work}
\vspace{-2pt}
\subsection{Legal Judgment Prediction}
\label{Sec:related_work_judge_predict}
Recently, legal judgment prediction has addressed much attention and achieved great progress. Thanks to the accessibility of the large amount of legal judgment data\footnote{European Court of Human Rights (ECHR): \url{https://echr.coe.int/Pages/home.aspx?p=home}; China Judgements Online: \url{http://wenshu.court.gov.cn/}}, a rising number of work has been dedicated to this research topic. However, they simplified the task to predict the court's outcome given a text describing the facts of a legal case manually summarized by the judge based on the case materials \cite{chalkidis-etal-2019-neural,ye2018interpretable,jiang2018interpretable,zhong2018legal,luo2017learning,hu2018few,zhong2020iteratively,xu2020distinguish,yang2019legal,chen2019charge,long2019automatic}. 
\citet{xiao2018cail2018} proposed to take the fact description as input and predict charges by using a criminal case dataset named CAIL in Chinese. \citet{zhong2018legal} constructed a topological network to capture the dependencies among subtasks via multi-task learning relying on the fact description as well.
\citet{chalkidis-etal-2019-neural} released the first large-scale English legal judgment prediction dataset. Compared to the previous work, we are the first to perform judicial decision making through court debate and pre-trial claim data where we comprehensively examine the case across various stages of the case life-cycle.

Table \ref{tab:related_work} compares the state-of-the-art works related to judgment prediction over different dimensions, including the task to be solved, case type, methodology (highlight) as well as the phases of dataset used in the study. It is clear that most of the previous work are conducted on stage $3$ where the judge-summarized fact summary is the algorithm input. To prove the effectiveness of our model in optimizing the case representation across different stages to challenge the idealization scenario, we compare our proposed model with four representative methods \cite{hu2018few}, \cite{xu2020distinguish}, \cite{chalkidis-etal-2019-neural} and \cite{long2019automatic}. \cite{hu2018few} represents the faction of using discriminative legal attributes for judgment prediction which emphasizes more on the judicial fairness during prediction. \cite{xu2020distinguish} stands for the research works based on distinguishing law articles for judgment prediction which has been proved to be effective especially for criminal cases. \cite{chalkidis-etal-2019-neural} leverages BERT to focus only on learning good representation of the pure input fact text for judgment prediction. \cite{long2019automatic} works also on the civil cases as ours and despite the fact summary, it also employs the claims of the case and the related law articles in the verdict as input. 

\vspace{-4pt}
\subsection{Multi-Task Learning}
The usage of multi-task learning models has become ubiquitous for many natural language processing tasks. For example, \citet{cao2017improving} leverages text classification to improve the performance of multi-document summarization via joint learning.
\citet{guo2018soft} improved abstractive text summarization with the
auxiliary tasks of question generation and
entailment generation. \citet{nishida-etal-2019-answering} used the QA
model for answer selection and extractive
summarization for evidence
extraction in a neural multi-task learning framework. In the field of legal judgment prediction, multi-task framework has been widely adopted for learning several target objectives simultaneously \cite{chalkidis-etal-2019-neural,jiang2018interpretable,luo2017learning,xu2020distinguish,zhong2018legal,yang2019legal}. In this work, we introduce a framework with a ``bionic design'' to conduct judicial admissibility inspection (main task) across different stages via the supervision of the recognized facts by the judge (auxiliary task).

\vspace{-4pt}
\section{Preliminaries}
\label{Sec:preliminary}
In this section, we first introduce the problem addressed in this paper. Then we provide an overview of the \textsf{MSJudge} framework to address it. The set of key notations used in this paper is given in Table~\ref{tab:notation}. Note that $U_i$, $r_i$, $S_i$, $w$ and $c$ represent the embedding representations of the corresponding variables in the table.
\vspace{-4pt}
\subsection{Problem Formulation}
Let $D=\left\{U_1, U_2, \cdots, U_n \right\}$ denote a court debate with $n$ utterances where each utterance $U_i$ consists of a sequence of $l$ words $S_i = \left\{ w_{i1},w_{i2},\cdots,w_{il} \right\}$ and the role of its speaker $r_i$. Each case has a set of $k$ claims $C=\left\{ c_1,c_2,\cdots,c_k \right\}$ where each claim is composed of a sequence of $q$ words $c_j=\left\{ w_{j1},w_{j2},\cdots,w_{jq} \right\}$. Based on the debate and claims of a case, the main task aims to predict judgment results $Y^c=\left\{y_1^c,y_2^c,\cdots,y_k^c\right\}$ of all claims. In order to imitate the judge's decision process, the model predicts $z$ fact labels  $Y^f=\left\{y_1^f,y_2^f,\cdots,y_z^f\right\}$ simultaneously as an auxiliary task to improve the results of the main task. 

\begin{table}[!t]
\caption{Definition of Notations}
\scriptsize
\vspace{-2pt}
\label{tab:notation}
\begin{tabular}{|cl|}
\hline
$D$ & a debate dialogue containing $n$ utterances \\
$U_i$ & the $i$-th utterance in $D$ \\
$S_i$ & the text content of $U_i$ \\
$r_i$ & \makecell[l]{the role of the speaker in $U_i$(i.e. judge, 
plaintiff, defendant and witness)} \\
$w_{it}^u$ & the $t$-th word in $S_i$ \\
$C$ & a set of claims of the case\\
$c_j$& the $j$-th claim in $C$\\
$w_{jv}^c$& the $v$-th word in $c_j$\\
$y_p^f$ & the predicted probability of $p$-th fact\\
$y_j^c$ & \makecell[l]{the predicted probability distribution of $j$-th claim}\\ 
\hline
\end{tabular}
\vspace{-12pt}
\end{table}

\vspace{-2pt}
\subsection{Overview}
As formulated above, in civil litigation, the judgment prediction is conducted on each claim, thus the proposed case representation is rooted in those claims (from stage 1) and intensified with the recognized facts from court debate (from stage 2). Modeling the interactions among claims, debate and recognized facts becomes essential in judgment prediction especially in case life-cycle scenario. 

\begin{figure*}[!t]
\centering
\includegraphics[width=0.8\linewidth]{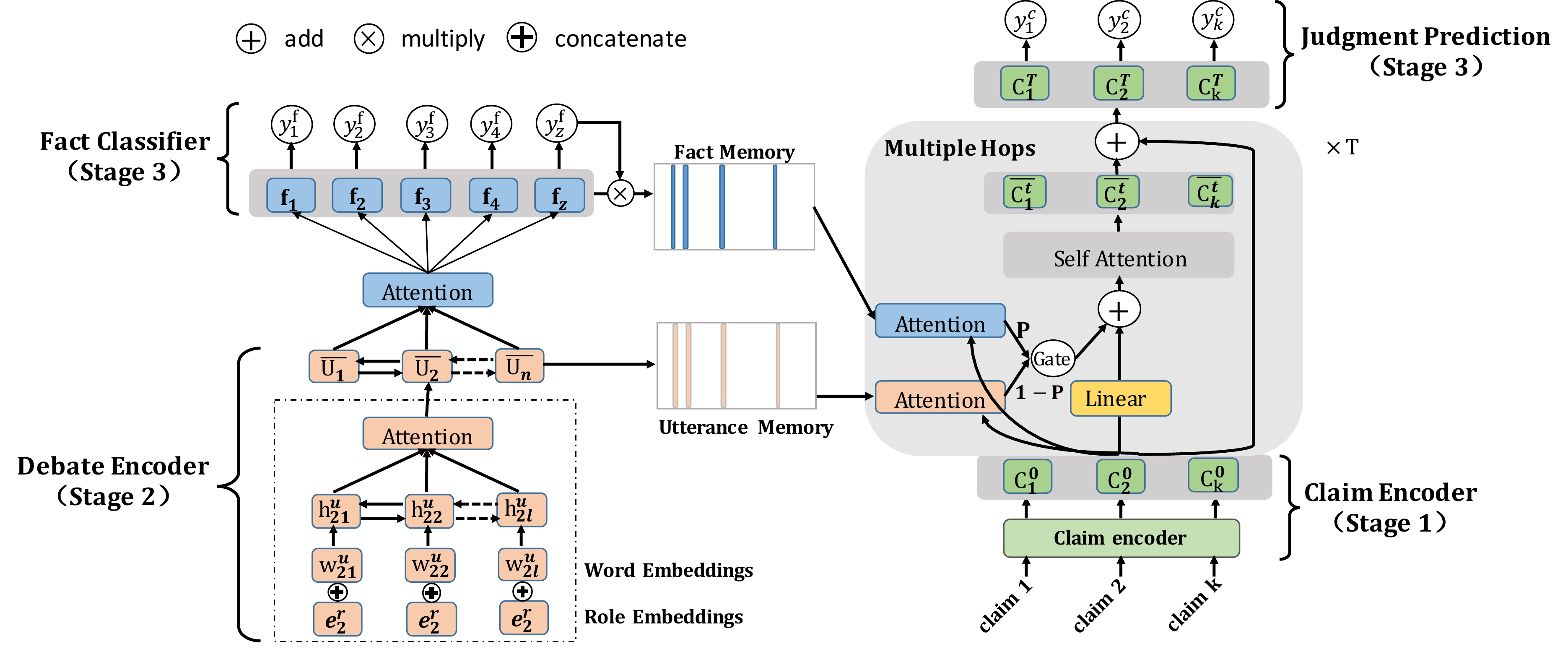}
\vspace{-4pt}
\caption{Architecture of MSJudge. The model is divided into three modules: (1) Multi-Stage Context Encoding: encoding claim and debate. (2) Multi-stage Content Interaction: capturing the correlations among claim, debate and fact. (3) Fact Recognition and Judgment Prediction: identifying the legal facts and then make judgment with respect to each claim.}
\vspace{-12pt}
\label{fig:model_description}
\end{figure*}

Fig.~\ref{fig:model_description} depicts the architecture of \textsf{MSJudge}. It learns multi-stage context representation and capturing the mutual effect across different stages of data to further make judgment with respect to the claims. Specifically, it is composed of the following three major components.

\begin{itemize}
\item \textit{Multi-Stage Context Encoding} simulates a judge to parse and understand the court debate and its pre-trial claims.  
\item \textit{Multi-Stage Content Interaction} simulates a judge's need to find clues and diagnose the correlations of the key information in the multi-stage content (e.g., claims, court debate). We model the interaction between utterances and claims, interaction between facts and claims, as well as the interaction across claims to enhance the claim representations.
\item \textit{Fact Recognition and Judgment Prediction} identifies the legal fact by considering all the information mined in the previous step and then make final judgment with respect to each claim. 
\end{itemize}
\vspace{-2pt}
\section{The MSJUDGE Framework}
\label{Sec:model}

\subsection{Multi-Stage Context Encoding}
\subsubsection{Debate Utterance Encoder}
Given an utterance $U_i$ with $l$ words $S_i=\left\{ w_{i1}^u,w_{i2}^u,\cdots,w_{il}^u \right\}$ and the role of its speaker $r_i\in R$, we first embed the words to vectors to obtain $\hat{S_i}=\left\{ \textup{w}_{i1}^u,\textup{w}_{i2}^u,\cdots, \textup{w}_{il}^u \right\}$ where $\textup{w}^u \in \mathbb{R}^d$ and employ role embedding to encode the role. The role embedding $\textup{e}_i^r \in \mathbb{R}^r$ is randomly initialized and jointly learnt during the training process.  

To involve the role information into the utterance, we concatenate the role information with each word in the utterance, which is able to project the same word into different dimensional spaces w.r.t.the target role. We hypothesize that the same word may need differentiate when different speakers use it.
\setlength{\abovedisplayskip}{2pt}
\setlength{\belowdisplayskip}{2pt}
\begin{equation}
\begin{aligned}
\textup{e}_{it}^u = \textup{w}_{it}^u \oplus \textup{e}_i^r,\ t\in[1, l]
\end{aligned}
\end{equation}

where $\oplus$ denotes a concatenation operation and then the dimention of $\textup{e}_{it}^u$ is $(d+r)$.

 Then we utilize a bidirectional-LSTM to encode the semantics of the utterance while maintaining its syntactic.
\setlength{\abovedisplayskip}{2pt}
\setlength{\belowdisplayskip}{2pt}
\begin{equation}
\label{equ:Bi-LSTM}
\begin{aligned}
\textup{h}_{it}^u &= [\overrightarrow{\textrm{LSTM}^{U}(\textup{e}_{it}^u)}, \overleftarrow{\textrm{LSTM}^{U}(\textup{e}_{it}^u)}],\ t\in[1, l]
\end{aligned}
\end{equation}

where $\textup{h}^u_{it}$ is the $t$-th word in $i$-th utterance's representation. 

To strengthen the relevance between words in an utterance, we employ the attention mechanism to obtain $\textup{U}_i$, which can be interpreted as a local representation of an utterance:
\setlength{\abovedisplayskip}{2pt}
\setlength{\belowdisplayskip}{2pt}
\begin{equation}
\begin{aligned}
\textup{U}_i &= \sum_{t=1}^{l} \alpha_{it}^u \textup{h}_{it}^u\\
\alpha_{it}^u &= \frac{\textup{exp}(\textup{Q}^u\textup{h}_{it}^u)}{\sum_{t=1}^{l} \textup{exp}(\textup{Q}^u\textup{h}_{it}^u)}
\end{aligned}
\end{equation}
 
where $\textup{Q}^u$ are learnable parameters and all parameters in utterance encoder are shared across utterances.
 
\vspace{-8pt}
\subsubsection{Debate Dialogue encoder}
To represent the global context in a dialogue, we use another bidirectional-LSTM to encode the dependency between utterances to obtain a global representation of an utterance, denoted as $\overline{\textup{U}_i}$.
\setlength{\abovedisplayskip}{2pt}
\setlength{\belowdisplayskip}{2pt}
\begin{equation}
\begin{aligned}
\overline{\textup{U}_i} &= [\overrightarrow{\textrm{LSTM}^{D}(\textup{U}_i)}, \overleftarrow{\textrm{LSTM}^{D}(\textup{U}_i)}],\ i\in[1, n]
\end{aligned}
\end{equation}

where $\overline{\textup{U}_i}$ is the $i$-th utterance's global representation.

\subsubsection{Pre-trial Claim Encoder}
Similar to the utterances, we encode the claims via bidirectional-LSTM and use attention mechanism to obtain the local representations of claims. 
We share word embedding matrix across the utterance encoder and the claim encoder.
\setlength{\abovedisplayskip}{2pt}
\setlength{\belowdisplayskip}{2pt}
\begin{equation}
\begin{aligned}
\textup{h}_{jv}^c &= [\overrightarrow{\textrm{LSTM}^{C}(\textup{w}_{jv}^c)},\overleftarrow{\textrm{LSTM}^{C}(\textup{w}_{jv}^c)}],\ v\in[1, q]\\
\textup{C}_j &= \sum_{v=1}^{q} \alpha_{jv}^c \textup{h}_{jv}^c\\
\alpha_{jv}^c &= \frac{\textup{exp}(\textup{Q}^c\textup{h}_{jv}^c)}{\sum_{v=1}^{q} \textup{exp}(\textup{Q}^c\textup{h}_{jv}^c)}
\end{aligned}
\end{equation}

where 
$\textup{C}_j$ is the $j$-th claim's representation and $\textup{Q}^c$ are learnable parameters and the parameters  are shared across claims.

\vspace{-4pt}
\subsection{Multi-Stage Content Interaction}
\label{Sec:interaction}
\subsubsection{Debate-to-Claim}
\label{Sec:debate-to-claim}
Utterance vectors are stacked and regarded as an utterance memory $\textup{m}^u = \left\{ \overline{\textup{U}_1},\overline{\textup{U}_2},\cdots,\overline{\textup{U}_n} \right\}$. We compute attention weights where each weight indicates the correlation between a claim vector $\textup{C}_j$ and an utterance memory unit $\textup{m}_i^u$.
\setlength{\abovedisplayskip}{2pt}
\setlength{\belowdisplayskip}{2pt}
\begin{equation}
\begin{aligned}
\textup{O}^u_j &= \sum_{i=1}^n \alpha_{ji}^d \overline{\textup{U}_i}\\
\alpha_{ji}^d &= \frac{\textup{exp}(\textup{C}_j \overline{\textup{U}_i})}{\sum_{i=1}^{n} \textup{exp}(\textup{C}_j \overline{\textup{U}_i})}
\end{aligned}
\end{equation}

where $\textup{O}^u_j$ is the output vector of the interaction between utterance memory and a claim.
\vspace{-4pt}
\subsubsection{Debate-to-Fact}
\label{Sec:debate-to-fact}
As introduced above, the identiﬁed judicial facts are the key factors for the judge to make decisions of the target case, thus we conduct judicial fact recognition as auxiliary task (see Sec. \ref{Sec:auxiliary_task}). To discover the projection of different facts on the certain dialogue fragments in a debate, we employ attention mechanism to obtain fact representation $\textup{f}_p$ for the $p$-th fact.
\setlength{\abovedisplayskip}{2pt}
\setlength{\belowdisplayskip}{2pt}
\begin{equation}
\begin{aligned}
\textup{f}_p &= \sum_{i=1}^n \alpha_{pi}^r \overline{\textup{U}_i}\\
\alpha_{pi}^r &= \frac{\textup{exp}(\textup{Q}_p^r \overline{\textup{U}_i})}{\sum_{i=1}^{n} \textup{exp}(\textup{Q}_p^r \overline{\textup{U}_i})}
\end{aligned}
\end{equation}
 
where $\textup{Q}_p^f$ are learnable parameters.
\vspace{-4pt}
\subsubsection{Fact-to-Claim}
\label{Sec:fact-to-claim}
Similar to utterance memory, the fact vectors are multiplied by the prediction probability and then 
we stack the fact vectors of the facts 
as a fact memory $\textup{m}^f = \left\{ \overline{\textup{f}_1},\overline{\textup{f}_2},\cdots,\overline{\textup{f}_z} \right\}$. 
Moreover, we attentively sum the fact memory to produce output representation.
\setlength{\abovedisplayskip}{2pt}
\setlength{\belowdisplayskip}{2pt}
\begin{equation}
\begin{aligned}
\textup{O}^f_j &= \sum_{p=1}^z \alpha_{jp}^f \overline{\textup{f}_p}\\
\alpha_{jp}^f &= \frac{\textup{exp}(\textup{C}_j \overline{\textup{f}_p})}{\sum_{p=1}^{z} \textup{exp}(\textup{C}_j \overline{\textup{f}_p})}
\end{aligned}
\end{equation}

where $\textup{O}^f_j$ is the output vector of the interaction between fact memory and a claim.
\vspace{-4pt}
\subsubsection{Fusion}
For each claim, we obtain the output $\textup{O}^u_j$ from utterance memory and the output $\textup{O}^f_j$ from fact memory. We use gate mechanism to control the weight of the two vectors to pass to the next layer. We further apply a linear layer with Rectifier Liner Unit (ReLU) to obtain $\hat{\textup{C}_j}$. After the addition, we get $\overline{\textup{C}_j}$ as the claim representation via memory blocks.
\setlength{\abovedisplayskip}{2pt}
\setlength{\belowdisplayskip}{2pt}
\begin{equation}
\begin{aligned}
\textup{g}_j &= \sigma(W^u\textup{O}^u_j+W^f\textup{O}^f_j+b^g)\\
\hat{\textup{C}_j} &= \textup{ReLU}(W^l\textup{C}_j+b^l)\\
\overline{\textup{C}_j} &= \hat{\textup{C}_j}+g_j*\textup{O}^u_j+(1-g_j)*\textup{O}^f_j
\end{aligned}
\end{equation}

where $\sigma$ is sigmoid activation function, $W^u$, $W^f$, $W^l$, $b^g$ and $b^l$ are trainable parameters shared across claims.
\vspace{-8pt}
\subsubsection{Across-Claim}
\vspace{-10pt}
\label{Sec:across_claim}
As aforementioned it is common to have multiple claims in one case and whether they are established or not is not relatively independent, it is necessary to model the dependency across claims. Technically, we employ self attention mechanism to capture the relationships across claims.
\setlength{\abovedisplayskip}{2pt}
\setlength{\belowdisplayskip}{2pt}
\begin{equation}
\begin{aligned}
Attention(Q,K,V) &= softmax(\frac{QK^T}{\sqrt{d_k}})V\\
\end{aligned}
\end{equation}

where $Q \in \mathbb{R}^{d_k \times k},K\in \mathbb{R}^{d_k \times k},V\in \mathbb{R}^{d_k \times k}$ are query, key, value which are the same vector. 

We take the stack of claim vectors $\overline{\textup{C}^c}=\left\{ \overline{\textup{C}_1},\overline{\textup{C}_2},\cdots,\overline{\textup{C}_k} \right\}$ as input to a self-attention layer with residual
connections.

Moreover, we employ multiple ($T$) hops (denoted as the grey block in Fig. \ref{fig:model_description}) in our model where the output of the previous hop is considered as the input of next hop. Previous works \cite{sukhbaatar2015end,tang2016aspect} have proved the usage of multiple hops in memory network could yield learn the deep abstraction of text.
\vspace{-12pt}
\subsection{Task-specific Decoders}
\subsubsection{Main Task: Judgment Prediction}
After $T$ hops updates, we obtain the final claim representation $\textup{C}_j^T$ for $j$-th claim and feed it to the softmax
layer for judgment prediction. 
\setlength{\abovedisplayskip}{2pt}
\setlength{\belowdisplayskip}{2pt}
\begin{equation}
\begin{aligned}
y_j^c&=\textup{softmax}(W^c\textup{C}_j^T+b^c)\\
\end{aligned}
\end{equation}
We train our model in an end-to-end manner by minimizing the cross-entropy loss.
\setlength{\abovedisplayskip}{2pt}
\setlength{\belowdisplayskip}{2pt}
\begin{equation}
\begin{aligned}
\mathcal{L}_c&=-\frac{1}{k}\sum_{j=1}^k\sum_{d=1}^{\left|Y_c\right|} g_{jd}^c \textup{log}(y_{jd}^c)\\
\end{aligned}
\end{equation}

where $g_{jd}^c$, $y_{jd}^c $ are the ground truth and the predicted probability of $d$-th class for $j$-th claim for each training instance, respectively.

\subsubsection{Auxiliary Task: Judicial Fact Recognition}
\label{Sec:auxiliary_task}
As for the judicial fact recognition task, we feed fact representation $\textup{f}_p$ to the sigmoid layer to predict the probability of a fact should be recognized.
\setlength{\abovedisplayskip}{2pt}
\setlength{\belowdisplayskip}{2pt}
\begin{equation}
\begin{aligned}
y_p^f &=\textup{sigmoid}(W_p^f \textup{f}_p+b_p^f)\\
\end{aligned}
\end{equation}

Similarly, the training loss is constructed as:
\setlength{\abovedisplayskip}{2pt}
\setlength{\belowdisplayskip}{2pt}
\begin{equation}
\begin{aligned}
\mathcal{L}_f&=-\frac{1}{z}\sum_{p=1}^z\left[g_{p}^f\textup{log}(y_{p}^f)+(1-g_{p}^f)\textup{log}(1-y_{p}^f)\right]\\
\end{aligned}
\end{equation}

where $g_{p}^f$, $y_{p}^f$ are the ground truth and the predicted fact recognition probability of the $p$-th fact for each training instance.

Finally the total loss is the sum of these two losses:
\setlength{\abovedisplayskip}{2pt}
\setlength{\belowdisplayskip}{2pt}
\begin{equation}
\begin{aligned}
\mathcal{L}_{total} &=\mathcal{L}_c+\mathcal{L}_f\\
\end{aligned}
\end{equation}

\vspace{-4pt}
\section{Experiment Settings}
\vspace{-2pt}
\subsection{Dataset  Construction}
In the experiment, we collected $70,482$ cases of Private Lending category. Each case includes plaintiff's claims, court debate records and judgment verdict. In total, it contains more than $4.1$ million utterances and $133,209$ claims. On average, each case contains $58.17$ utterances and $1.89$ claims and the average number of words of debates is 950 compared to the average number of words of fact summary is 143. The ratio of category labels\footnote{The labels of main task classifier: \textit{reject}, \textit{partially support} and \textit{support}} in main task is $1:2.6:10.9$. We define $10$ fact labels in auxiliary task\footnote{The labels of auxiliary task classifier: 
\textit{Agreed Loan Period}, \textit{Couple Debt}, \textit{Limitation of Action}
, \textit{Liquidated Damages}, \textit{Repayment Behavior}, \textit{Term of Guarantee}, \textit{Guarantee Liability}, \textit{Term of Repayment}, \textit{Interest Dispute}, \textit{Loan Established}}. The labeling of the fact label is based on the fact finding section in the verdict, and the judgment result is annotated based on the judgment paragraph in the verdict. All the annotations are conducted by legal experts with three rounds of training led by civil judges. In the end, the experts achieved Kappa coefficient = 0.8 (substantial agreement).
To the best of our knowledge, this is the very first large aligned civil trial court debate and judgment document (case life-cycle) dataset. We show two complete and real life-cycle materials of the cases after removing sensitive information\footnote{https://github.com/mly-nlp/LJP-MSJudge} (claims, court records, and their verdicts) to expose how the judges hear the case and make judgment in practical situation. We will release all the experiment data to motivate other scholars to further investigate this problem.

\vspace{-8pt}
\subsection{Training Details}
The dimensions of word embeddings and role embeddings are set to $300$. Word embeddings are trained using the Skip-Gram model\cite{mikolov2013distributed} on the debate dialogues and role embeddings are randomly initialized. We tune the performance by following the grid search tuning method and cross-validation. The size of hidden states of bidirectional-LSTM is $256$. The neural networks
are trained using Adam Optimization\cite{kingma:adam} with a learning rate set to $0.001$, and perform the mini-batch gradient descent with a batch size of $16$. The dropout is set to $0.8$.

To minimize judicial discrimination, we preprocess the data and replace all the personal information (e.g., person name, ID number, address, and gender) with special characters before the model training. During the experiment, we adopted cross-validation to ensure the rationality of the models. We will release our code as well as the dataset for reproducibility.

\vspace{-8pt}
\subsection{Evaluation Metrics}
We use Macro F$_1$ and Micro F$_1$ (Mac.F$_1$ and Mic.F$_1$ for short) as the main metrics for algorithm evaluation. 
In a multi-class classification setup, macro-average reflects the robustness of the model if there exists class imbalance. 
Note that as for all the baselines, we set the debate content concatenated with each claim of the case as input\footnote{If a case contains $k$ claims, then such case forms to $k$ samples in which each sample input is the combination of the debate content and one of the $k$ claims.} and the judgment result for each claim as output. As for the proposed methods, we are capable of predicting the judgment results of all the claims of a case at once, thus in each sample we conduct $k$ multi-class classification tasks where $k$ is the number of claims in a case. 
\vspace{-8pt}
\subsection{Tested Methods}
\subsubsection{Baselines}
To extensively validate the effectiveness of the proposed model, the following baselines under different scenarios are employed for comparison. 

We first testify several variants of encoders by setting same input type as MSJudge using claim and court debate from stage 1 and 2:
\begin{itemize}[itemsep=2pt,topsep=0pt,leftmargin=*]
\item \textit{Traditional machine learning based method}

\textbf{TFIDF+SVM} is a robust multi-class classification by means of TFIDF plus SVM \cite{suykens1999least}.

\item \textit{Deep learning based methods}

\textbf{TextCNN} is a convolutional neural networks trained on top of pre-trained word vectors\footnote{Skip-gram model\cite{mikolov2013distributed} is utilized for pretraining word representations.} for sentence-level classification tasks  \cite{kim2014convolutional}.
\textbf{BiGRU-ATT} employs Bi-directional GRU with attention mechanism \cite{xu2015show} to capture context semantics and automatically selects important features through attention during training. For \textbf{TextCNN} and \textbf{BiGRU-ATT}, we use entire debate content as input.
\textbf{HAN} stands for Hierarchical Attention Network \cite{yang2016hierarchical} which is a hierarchical text classification model with two levels of attention mechanisms for aggregating words to utterance and utterances to dialogue.
\textbf{Transformer}\cite{vaswani2017attention} is based solely on attention mechanisms. We use the encoder part of the transformer encoding entire debate content, followed by classifiers.

\item \textit{Multi-task learning methods}

To validate the superiority of joint learning with auxiliary task (judicial fact recognition), we conduct experiments for the above models with the multi-task framework. 
These models are denoted as \textbf{TextCNN-MTL}, \textbf{BiGRU-ATT-MTL}, \textbf{Transformer-MTL} and \textbf{HAN-MTL}.
\end{itemize}

\smallskip
\noindent We then compare several state-of-the-art (SOTA) methods on judgment prediction.
\begin{itemize}[itemsep=2pt,topsep=0pt,leftmargin=*]

\item \textit{Discriminative fact attributes based method}

\textbf{Few-Shot(fact)}\footnote{\url{https://github.com/thunlp/attribute_charge}} \cite{hu2018few} proposes to discriminate confusing charge, which extracts predefined attributes from fact descriptions to enforce semantic information. In the experiment, we use the ten fact labels in the auxiliary task of a case. 

\item \textit{Law articles based method}

\textbf{LADAN(fact)}\footnote{\url{https://github.com/prometheusXN/LADAN}} \cite{xu2020distinguish} introduces a novel graph distillation operator (GDO) to extract discriminative features for distinguishing confusing law articles. The original work is in a multi-task framework in criminal cases. In our civil case settings, it is trained by the two objectives: judgment prediction and law prediction. 

\item \textit{Content based method}

\textbf{BERT(fact)}\footnote{\url{https://github.com/google-research/bert}} \cite{devlin2019bert} is a fine-tuning representation model which has been applied to learn good representation of the input fact summary for judgment prediction
\cite{chalkidis-etal-2019-neural}. In the experiment, we take the representation of ``[CLS]'' as aggregated representation and add a softmax layer on the top of BERT for judgment prediction.

\textbf{Autojudge(fact)} \cite{long2019automatic} formalizes the task of judgment prediction as a reading comprehension task where the claims, fact and the related law articles are injected as input. For fair comparison, the input of the referenced law articles is removed to avoid information leakage during judgment prediction.
\end{itemize}

We also evaluate their models by replacing the fact summary with the case life-cycle data (claims, debate and fact labels) as conducted in our settings (the rows \textbf{Few-Shot-MTL}, \textbf{LADAN-MTL} and \textbf{Autojudge-MTL} in Table \ref{tab:overall_performance}). The results, then, can comprehensively validate the methodological and data hypotheses.

Note that all the baselines tested under the real court setting (see Table \ref{tab:overall_performance}) shared the same claim and debate encoder as MSJudge to make fair comparison.
\vspace{-6pt}
\subsubsection{Variants of our proposed method}
To comprehensively evaluate each component of the proposed method, we make several variants to address the corresponding issue.

\textbf{MSJudge-MTL} is our proposed method in multi-task framework. \textbf{MSJudge} is the single task version by removing the entire fact related parts (auxiliary task and fact memory). 
\textbf{MSJudge(fact)} takes the judge-summarized fact at stage 3 as input.

\vspace{-10pt}
\section{Result Discussion}
\label{Sec:result}
\begin{table}[!t]
\centering
\scriptsize
\caption{Main Results of All Tested Methods for the Main Task. Note that the average scores shown at rows \textbf{MSJudge(fact)}, \textbf{MSJudge} and \textbf{MSJudge-MTL} are statistically significant different from the corresponding value of all the baseline models (* denotes the $p$-value<$0.05$, $\dag$ denotes the $p$-value<$0.01$). \crule[yellow]{0.3cm}{0.3cm} \crule[green]{0.3cm}{0.3cm} \crule[blue]{0.3cm}{0.3cm} stand for employing the data at stage 1 (claim), stage 2 (court debate) and stage 3 (fact summary), 
respectively. \crule[gray]{0.3cm}{0.3cm} represents to employ the fact labels at stage 3 for supervision in training phase.}
\vspace{-2pt}
\label{tab:overall_performance}
\begin{tabular}{p{19mm}|p{18mm}|p{3mm}p{3mm}p{4mm}p{4mm}|p{0.5mm}p{0.5mm}p{0.5mm}|}
\toprule
&
\multicolumn{1}{c|}{\multirow{2}{*}{\textbf{Method}}} &
\multicolumn{4}{c|}{\textbf{Average}}&
\multicolumn{3}{c}{\textbf{Stage}}\\ 
\cline{3-9}
& & Mac.P &Mac.R &Mac.F1 &Mic.F1 &1&2&3\\
\midrule
\multirow{5}{*}{Pseudo setting}&Few-Shot(fact)\cite{hu2018few}  &76.3 &69.6 &72.5 &84.2&\cellcolor{yellow}&\cellcolor{white}&\cellcolor{blue} \\
&LADAN(fact)\cite{xu2020distinguish}  &76.8 &67.4 &71.2 &83.7&\cellcolor{yellow}&\cellcolor{white}&\cellcolor{blue} \\
&Autojudge(fact)\cite{long2019automatic}  &76.1 &69.4 &72.3 &83.9&\cellcolor{yellow}&\cellcolor{white}&\cellcolor{blue} \\
&BERT(fact) &76.5 &72.9 &74.6 &84.2&\cellcolor{yellow}&\cellcolor{white}&\cellcolor{blue} \\
&\textbf{MSJudge (fact)}   &\textbf{77.7*}& \textbf{73.4*} &\textbf{75.8*} &\textbf{86.6*}&\cellcolor{yellow}&\cellcolor{white}&\cellcolor{blue} \\\midrule

\midrule
\multirow{6}{*}{Real court setting} & TFIDF+SVM  &72.6 &53.6 &58.7 &79.9&\cellcolor{yellow}&\cellcolor{green}&\cellcolor{white}\\
& TextCNN &70.9 &62.0 &65.5 &81.1 &\cellcolor{yellow}&\cellcolor{green}&\cellcolor{white}\\
&Transformer &73.9 &63.8 &67.7 &82.0&\cellcolor{yellow}&\cellcolor{green}&\cellcolor{white} \\
&BiGRU-ATT &75.5 &66.7 &70.3 &83.1&\cellcolor{yellow}&\cellcolor{green}&\cellcolor{white} \\
&HAN &75.8 &67.0 &70.4 &83.2 &\cellcolor{yellow}&\cellcolor{green}&\cellcolor{white}\\
&\textbf{MSJudge}   &\textbf{76.3*}& \textbf{71.4*} &\textbf{73.6*} &\textbf{84.6*}&\cellcolor{yellow}&\cellcolor{green}&\cellcolor{white} \\
\midrule
\multirow{8}{*}{Real court setting-MTL} &TextCNN-MTL&70.8 &62.8 &66.1 &81.4&\cellcolor{yellow}&\cellcolor{green}&\cellcolor{gray} \\
&BiGRU-ATT-MTL &75.5 &68.0 &71.2 &83.3&\cellcolor{yellow}&\cellcolor{green}&\cellcolor{gray} \\
&Transformer-MTL &74.8 &68.9 &71.5 &82.2&\cellcolor{yellow}&\cellcolor{green}&\cellcolor{gray} \\
&HAN-MTL  &75.7 &69.0 &71.8 &83.1&\cellcolor{yellow}&\cellcolor{green}&\cellcolor{gray} \\ 
&Few-Shot-MTL\cite{hu2018few} &76.1 &69.6 &72.3 &83.5&\cellcolor{yellow}&\cellcolor{green}&\cellcolor{gray} \\
&LADAN-MTL\cite{xu2020distinguish} &74.5 &66.6 &69.8 &83.1&\cellcolor{yellow}&\cellcolor{green}&\cellcolor{gray} \\
&Autojudge-MTL\cite{long2019automatic}  &72.5 &71.8 &72.2 &83.2 &\cellcolor{yellow}&\cellcolor{green}&\cellcolor{gray} \\ 
&\textbf{MSJudge-MTL} &\textbf{77.5$\dag$} &\textbf{72.5$\dag$} &\textbf{74.8$\dag$} &\textbf{86.5$\dag$}&\cellcolor{yellow}&\cellcolor{green}&\cellcolor{gray}\\ 
\bottomrule 
\end{tabular}
\vspace{-12pt}
\end{table}

\vspace{-4pt}
\subsection{Overall Performance}
To evaluate the performance of the proposed model, we export the results from the following perspectives:

\textbf{Comparison against the baselines in real court setting.}
Table \ref{tab:overall_performance} summarizes the performance of all the tested methods for the main task. The column ``\textit{Stage}'' indicates the case life cycle segment(s) used for prediction. Based on the results, the following observations are recorded: (1) In real court setting group, it is not surprising to see that the traditional machine learning based methods didn't perform well in terms of F$_1$ score. It indicates the importance of legal case representation learning for better judgment prediction. Among the deep learning based baselines, \textbf{HAN} outperforms the other ``single-level'' models for both macro F$_1$ and micro F$_1$ scores which indicates the necessity of using hierarchical context representation to capture the dependency within words, utterance, and dialogue in the court debate scenario.
(2) With the real court settings, we also employ multi-task training process (judgment prediction and judicial fact recognition) (see group Real court setting-MTL) for  \textbf{TextCNN-MTL}, \textbf{BiGRU-ATT-MTL}, \textbf{Transformer-MTL} and \textbf{HAN-MTL}, all of them show positive effects compared to their single task mode, which proofs the validity of employing multi-task training process in the judgment prediction task. Compared with SOTA methods (\textbf{Few-Shot-MTL}, \textbf{LADAN-MTL}, \textbf{Autojudge-MTL}), Since \textsf{MSJudge-MTL} is purposely designed to jointly learn the tasks of judgment prediction and fact recognition together. It outperforms all the other tested methods under case life-cycle scenario. 
Another finding worth to mention is that \textbf{LADAN-MTL} does not perform well in civil case scenario in that the law articles applied in criminal cases indicate the corresponding criminal charge but in civil cases the law articles are usually for reference use.

\textbf{Comparison with upper bound in pseudo setting}
In the experiment, we also let the proposed approach challenge the upper bound conducted in pseudo setting, i.e., using judge summarized case fact as input (see Group pseudo setting as shown in Table \ref{tab:overall_performance}). The proposed model with fact summary as input outperforms the state-of-the-art baselines which indicates the interactions between claims and fact summary as well as the interactions across the claims can be also effective in the idealization scenario. Meanwhile, the equivalent prediction performance between \textbf{MSJudge(fact)} and \textbf{MSJudge-MTL} grants us confidence that the proposed model is able to optimize the case representation across different stages.

Various types of inputs, either court debate data or fact summary (see Table \ref{tab:judicial fact recognition}), are compared by leveraging the results of judicial fact recognition (the auxiliary task). As aforementioned, because of content/logic gaps between the two stages' data, the judicial fact recognition results tell the similar pattern. By exploring the bad cases, one can, sometimes, highlight misalignments between court debate and judge summarized case fact, e.g., some evidence and material (from fact) are not mentioned in the court debate.

\begin{table}[!t]
\centering
\scriptsize
\caption{The comparison on the performance of judicial fact recognition with different input types.}
\vspace{-2pt}
\label{tab:judicial fact recognition}
\begin{tabular}{l|cc|cc}
\hline
\multirow{2}{*}{\diagbox[width=10em]{\textbf{Fact Label}}{\textbf{Input Type}}} & \multicolumn{2}{c|}{\textbf{Court Debate}} & \multicolumn{2}{c}{\textbf{Fact Summary}} \\
                      & \textit{Mic.$F_1$}         & \textit{Mac.F$_1$}        & \textit{Mic.F$_1$}         & \textit{Mac.F$_1$}        \\ \hline
Agreed Loan Period    & 73.1            & 70.6           & 83              & 81.9           \\
Couple Debt           & 96.8            & 59.7           & 92.3            & 75.1           \\
Limitation of Action  & 82.2            & 70.1           & 95.3            & 93.7           \\
Liquidated Damages    & 90.1            & 75.3           & 95.9            & 91.3           \\
Repayment Behavior    & 76.4            & 71.5           & 87.5            & 85.6           \\
Term of Guarantee     & 93.2            & 67.7           & 98.5            & 94.6           \\
Guarantee Liability   & 90.3            & 77.6           & 94.9            & 88.1           \\
Term of Repayment     & 84.2            & 55.6           & 83              & 81.9           \\
Interest Dispute      & 79.1            & 79.1           & 93.3            & 93.3           \\
Loan Established      & 90.5            & 79.8           & 96.2            & 92.4           \\ \hline
Averaged              & 85.6            & 83.7           & 93              & 92.3           \\ \hline
\end{tabular}
\vspace{-8pt}
\end{table}

\textbf{Component Assessment via Ablation Test.}
To assess the contribution of different components in the proposed method, we conduct ablation tests in Table 5.  To validate the influence of the interaction between fact and claim as illustrated in Sec. \ref{Sec:fact-to-claim}, we remove the attention layer between fact memory and claim (see the row ``w/o fact memory'')\footnote{Note that in this setting, the fact classifier still exist for regularizing the debate encoder during supervision.}. Similarly, the influence of utterance on representing claims and the impact across the claims are demonstrated in the rows ``w/o utterance memory'' and ``w/o self attention'' respectively. Table \ref{tab:Ablation Analysis table} clearly tells that all the components contribute positively to the results. To be specific, the \textit{utterance memory} feature has largest impact - their removal causes $22.6\%$ relative increase in error (RIE) for macro F$_1$ score. The interaction across claims shows significant impact for judgment prediction which demonstrates the claims are highly correlated. In addition, the \textit{role} information also has great influence on the judgment prediction tasks over dialogues. Experimental results prove that multi-stage case representation learning can be critical for legal AI system.

\begin{table}[!t]
\scriptsize
\centering
\begin{minipage}[b]{85mm}
    \begin{minipage}[t]{0.55\textwidth}
    \caption{Ablation Test.}
    \label{tab:Ablation Analysis table}
    \begin{tabular}{p{20mm}p{4mm}p{4mm}p{4mm}}
    \toprule  
    Models & Mic.F1 &Mac.F1 &RIE(\%)\\
    \midrule  
    MSJudge-MTL            &86.5 &74.8 & --\\
    \midrule
    w/o role embeddings      &85.2 &74.0 &3.2\\
    \midrule
    w/o utterance memory &82.8&69.1 &22.6\\
    \midrule
    w/o fact memory      &84.9 &74.6 &0.79\\
    \midrule
    w/o self attention   &83.6&72.5 &9.1 \\
    \bottomrule 
    \end{tabular}
    \vspace{-8pt}
    \end{minipage}
    \begin{minipage}[t]{0.45\textwidth}
    \caption{Effects of Hops}
    \label{tab:Effects of Multiple Hops}
    \begin{tabular}{p{10mm}p{8mm}p{8mm}}
    \toprule  
    \#Hops & Micro F1 &Macro F1\\
    \midrule  
    hops(1)   &85.2 &74.3\\
    \midrule
    hops(2)   &85.4 &74.5\\
    \midrule
    \textbf{hops(3)}   &\textbf{86.5} &\textbf{74.8}\\
    \midrule
    hops(4)   &85.0 &73.8\\
    \midrule
    hops(5)   &84.8 &73.3\\
    \midrule
    hops(6)   &84.3 &73.1\\
    \bottomrule 
    \end{tabular}
    \vspace{-8pt}
    \end{minipage}
\end{minipage}
\vspace{-8pt}
\end{table}

\vspace{-8pt}
\subsection{Convergence Analysis}
\begin{figure}[!t]
\vspace{-8pt}
\scriptsize
\centering
\begin{minipage}[b]{85mm}
    \subfloat{
        \begin{minipage}[t]{0.5\textwidth}
        \centering
        \includegraphics[width=1\linewidth,height=0.68\linewidth]{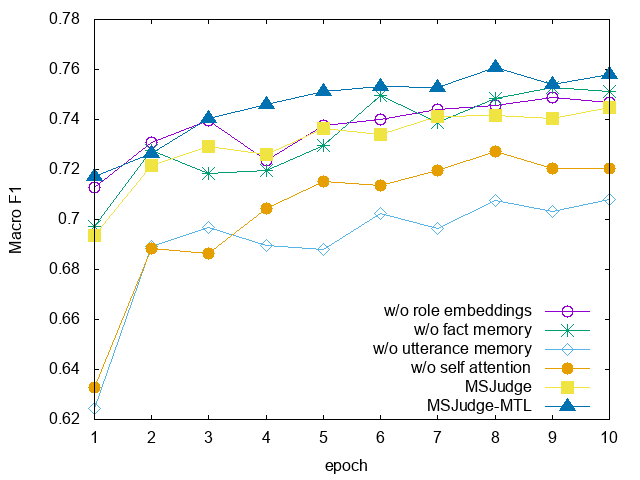}
        \caption{The performance of tested methods over epochs.}
        \label{fig:convergence}
        \vspace{-12pt}
        \end{minipage}
    }    
    \subfloat{
        \begin{minipage}[t]{0.5\textwidth}
        \centering
        \includegraphics[width=\linewidth,height=.68\linewidth]{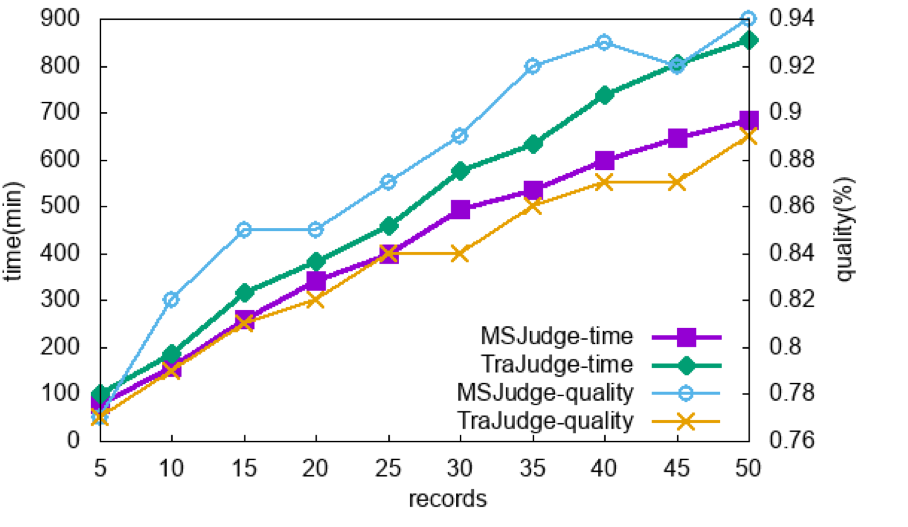}
        \vspace{-12pt}
        \caption{User Study: Judgment Efficiency and Quality.}
        \label{fig:user_study_plot}
        \end{minipage}
    }
\end{minipage}
\end{figure}

To further validate the performance of the proposed model, we conduct convergence analysis to monitor the changes and trends during training for different variants of the proposed model. As Fig. \ref{fig:convergence} depicts, one can observe that the performance of the model with training on all components is consistently superior than the models removing a particular component.

\begin{figure}[!t]
\centering
\includegraphics[width=1\linewidth,height=0.55\linewidth]{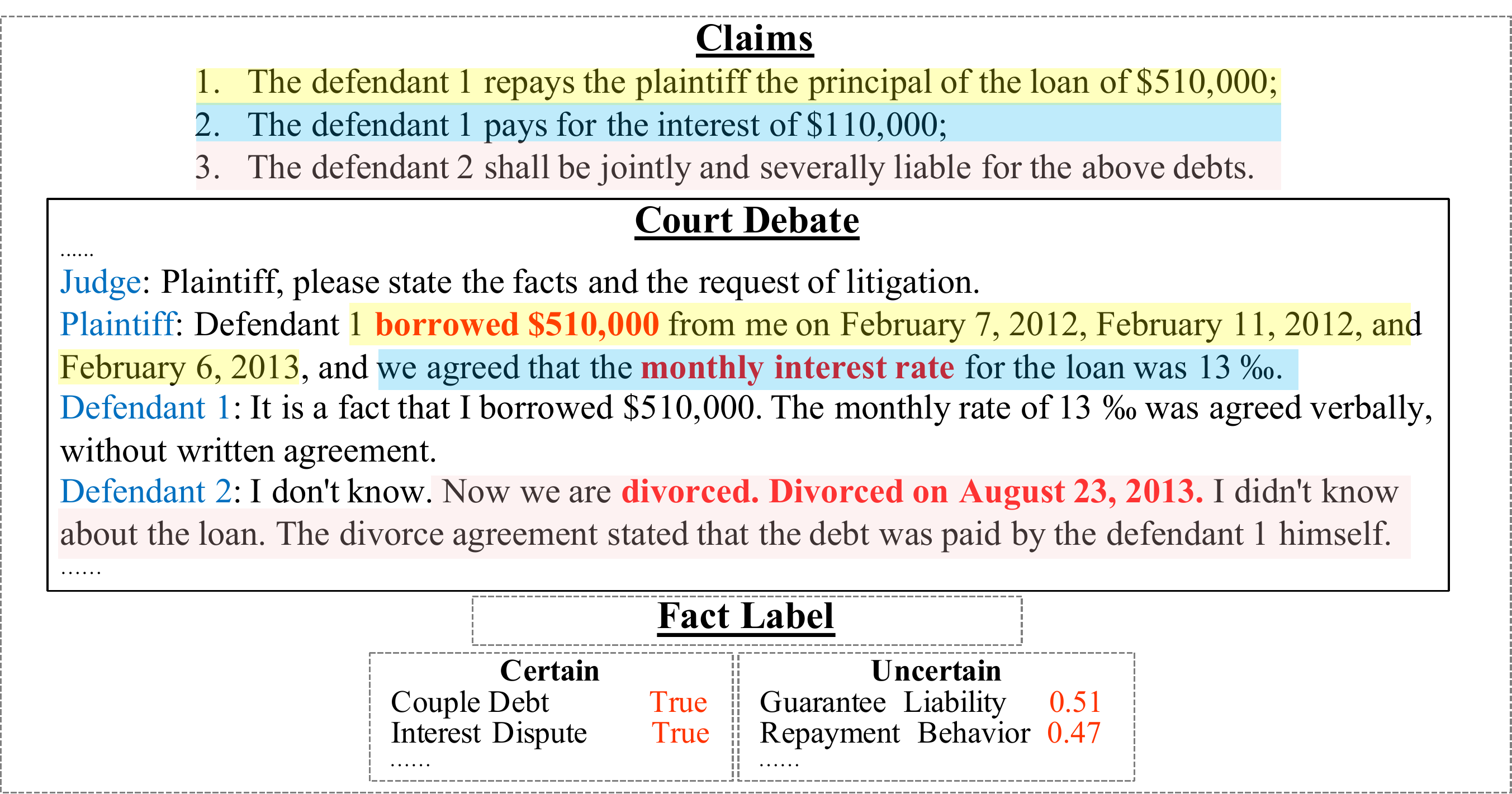}
\vspace{-12pt}
\caption{User Study: example of context enhanced by \textsf{MSJudge}.}
\label{fig:user_study_interface}
\vspace{-12pt}
\end{figure}

\vspace{-10pt}
\subsection{Effect of Multiple Hops}
\vspace{-4pt}
Table \ref{tab:Effects of Multiple Hops} shows the performances of our model with $1$ to $6$ memory hops where MSJudge-MTL(t) means our model using $t$ memory hops. The results show that our model with $3$ memory hops achieves the best result and then starts decreasing as the number of hops increases, which might be due to the loss of generality with the increase of model complexity\cite{tang2016aspect,zhu2018enhanced,xu2019multi}. Note that the parameters are shared over hops, thus the amount of parameters will not increase as the number of hops increases. We use the parameters at best hop for experimental results and visualization in Fig. \ref{fig:case_study}.

\vspace{-10pt}
\subsection{Model Interpretability}
\begin{figure*}[!t]
\centering
\includegraphics[width=.72\linewidth]{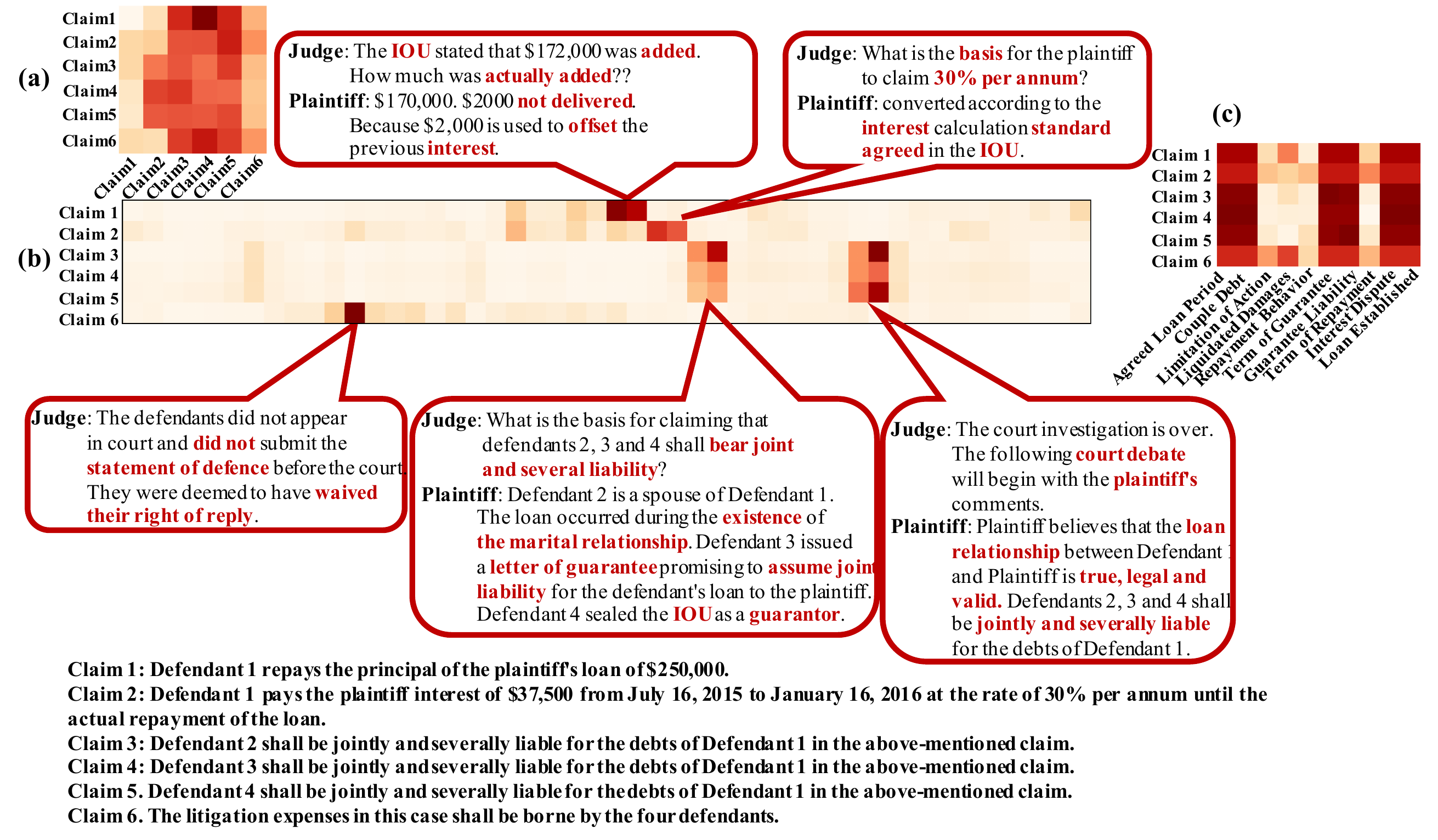}
\vspace{-8pt}
\caption{Model Interpretability. (a) depicts the dependency across claims; (b) describes the importance of the utterances/their key phrases to the claims; and (c) shows the correlation between facts and claims.}
\label{fig:case_study}
\vspace{-8pt}
\end{figure*}

\vspace{-4pt}
To help readers better consume our algorithm outcomes, Fig. \ref{fig:case_study} depicts a case study by visualizing all intermediate results at multiple stages for interpretations. In this case, the plaintiff raised in total $6$ claims to initialize the case where the claims $3,4,5$ get judicial support and the rest are partially supported. Our predictions for these claims match the ground truth perfectly. Fig. \ref{fig:case_study}(a) depicts the dependency across claims which is plotted by the weights on the self-attention layer as described in Section \ref{Sec:across_claim}. In this case we can observe claims $3,4,5$ have similar patterns due to their common petition of joint and several liability for the three guarantors. Similarly, Fig. \ref{fig:case_study}(c) visualizes the correlation between claims and fact labels\footnote{The weights on fact memory block.}, where claims $3,4,5$ were demonstrated to be highly correlated to almost the same set of fact labels about guarantee. Fig. \ref{fig:case_study}(b) shows the importance of the utterances to the corresponding claims\footnote{The weights on utterance memory block where the darker the color, the more important a certain utterance is to a target claim.} as well as the key phrases highlighted within each utterance\footnote{The value on the attention layer of debate encoder.}.
\vspace{-8pt}
\subsection{User Study}
\label{Sec:user_study}
\vspace{-4pt}
To justify the role of MSJudge in assisting the judges to make judicial decisions in real court setting, we conduct a user study by exploring the judgment effect with/without \textsf{MSJudge}. We select $12$ judges as subjects who were either judges or law school students. They are from three age groups ($18$-$28$, $29$-$40$, $41$-$65$) with four people in each group, where half of them use \textsf{MSJudge} (denoted as \textit{\textsf{MSJudge} Judge}), and the other half do not (denoted as \textit{Traditional Judge}).

\begin{table}[!t]
\scriptsize
\centering
\caption{User Study:open survey on MSJudge based on age}
\label{tab:User Study:open survey}
\begin{tabular}{c|c|c|c}
\hline 
\textbf{Questions} & \textbf{18-28} &\textbf{29-40} &\textbf{41-65}\\
\hline   
MSJudge can be helpful in fact prediction. &4.3 &4.1 &3.9\\
\hline 
MSJudge can capture the key information.  &3.8 &3.8&4.1\\
\hline
MSJudge can be used in real court setting. &3.8 &4.1&3.3\\
\hline 
MSJudge can help reduce the workload of trial judges.  &3.7 &4.0&3.4\\
\hline  
\end{tabular}
\vspace{-12pt}
\end{table}

We randomly selected $50$ real cases\footnote{Each case contains the claims from plaintiff and the corresponding court debate transcript, which performs the same setting with the input of the proposed MSJudge model.} and let the subjects to make judicial decisions. For each case, the subject should provide the legal fact recognized in this case and the final judgment to each claim. The judgment effect was evaluated after they finish every $5$ cases.
The \textsf{MSJudge} facilitates the decision making by automatically highlighting the most relevant keyword or sentence in the court debate with respect to each claim according to debate-to-claim attention as described in Section \ref{Sec:debate-to-claim}. The \textit{\textsf{MSJudge} judge} can then quickly capture the key information in the court debate for making judgment, rather than go over all the context. In addition, The \textsf{MSJudge} displays the predicted legal facts with their certainty/uncertainty as illustrated in Section \ref{Sec:auxiliary_task}. We define those facts with a probability greater than $0.7$ as certain facts, and those with a probability between $0.45$ and $0.55$ as uncertain facts. Thus those uncertain ones warn the subjects to focus on the potential controversies before making judgment. An example of enhanced context with predicted facts is shown in Fig.\ref{fig:user_study_interface}. 
The \textit{Traditional judge} instead need to read through the entire court debate and analyze the logic by themselves. Note that all the subjects enabled to check the correct answers (the ground truth of legal facts and judgments) once they finish every $5$ cases.

We evaluate the judgment effect of the two approaches from two perspectives:
(a) \textbf{Judgment Efficiency} measures the time the subject consuming on analyzing the case till making judicial decisions. The lesser the time cost, the more efficient).
(b) \textbf{Judgement quality} estimates whether the subject can correctly identify legal facts, and at the same time make correct judgments based on the identified legal facts. Thus the judgment quality was formalized as the weighted average of the accuracy of legal fact recognition and the accuracy of judgment to each claim.

Fig. \ref{fig:user_study_plot} depicts the change curves of two groups in terms of consuming time and judgment quality, with respect to the increase of the testing cases. It is worth mentioning that as the number of case increases, the growth rate of the consuming time of \textsf{MSJudge} (the curve \textsf{MSJudge}-time) tends to decline in contrast to the curve TraJudge-time. In other words, the learning efficiency gradually increases for \textsf{MSJudge}. Moreover, the \textit{\textsf{MSJudge} subjects} are found to achieve higher judgment quality (see curve MSJudge-quality) compared to the \textit{traditional subjects} (see curve TraJudge-quality).

In addition, we surveyed the \textit{\textsf{MSJudge} subjects} with several open questions as shown in Table \ref{tab:User Study:open survey}, to assess the opinions on the usage of MSJudge in real court setting. For each question, Likert scale of $1$ to $5$ needs to be selected; the higher the score, the more agreement to a statement.
Based on the survey results, we found that most people think \textsf{MSJudge} can be helpful in identifying legal facts from colloquial and lengthy court debate, and it accelerates the process of locating key information for making final judgment. As for the practical use of MSJudge in real court to reduce the workload of judges, the age range in the $18$-$28$ and $29$-$40$ responded positively, while the $41$-$65$ group had some reservations. But they still thought it is possible to involve the active supervision of a judge while using \textsf{MSJudge} as assistance in the real court.
\vspace{-10pt}
\subsection{Error analysis}
\vspace{-4pt}
For the bad cases, 12\% of the errors come from the fact identification when the court debate does not contain the utterances focusing on the corresponding fact which might be discovered only in the evidences (those data are not available). In such context, the evidence analysis can be another promising future work for the case life-cycle learning. Meanwhile, according to the confusion matrix, we also find that the specificity of the semantics in the court debate impairs the model performance when discriminating the labels of ``partially support'' and ``support''.
\vspace{-10pt}
\subsection{Ethical Statement}\label{section:ethical}
\vspace{-4pt}
Lastly, we would like to discuss ethical concerns of our work. We are aware of potential risks of structured social biases that can be repeated or even enhanced in any machine learning system trained on large-scale uncontrolled datasets \cite{bias2-onlineInfo}. The need to assure equality, judicial impartiality, and judicial diversity must be properly addressed, and is a topic of great significance in judicial decision making. 
To combat these concerns, we anonymized the data by removing sensitive information (e.g., gender, race, etc.). We also applied the technique of oversampling to cope with infrequent questions to be generated to assure well-balanced training dataset. 
Despite the potential bias, the potential system error would be as follows:
a) recognizing a wrong legal fact and b) generating a wrong judgment result. As for these concerns, in the user study, we managed to provide the users with the certainty/uncertainty of the identified facts to show some warnings for those uncertain facts. In addition, we suggest the future automatic judgment prediction system should be under Man-machine inclusive mode, which allows users to make corrections at the key points (e.g., legal fact identification) before the last step (e.g., judgment prediction) to ensure the correctness of the final conclusion. 
And indeed, when we tried to manually correct the answer at fact identification step, the performance judgment prediction increases significantly. 

\vspace{-10pt}
\section{Conclusion}
\vspace{-4pt}
Performing case life-cycle admissibility inspection over court debate can be practically useful to assist the judges to adjudicate cases. In this work, we introduce a novel and challenging dataset addressing the life-cycle case representation technique. An end-to-end framework \textsf{MSJudge} is proposed which is in a manner of multi-task learning process. The empirical findings validate our hypothesis that joint learning with auxiliary task can improve the performance over state-of-the-art approaches. Additionally, MSJudge can more accurately characterize the interactions among claims, fact and debate for judgment prediction, achieving significant improvements over strong state-of-the-art baselines. Moreover, the user study conducted with judges and law school students shows the neural predictions can also be interpretable and easily observed, and thus enhancing the trial efficiency and judgment quality. 
\vspace{-10pt}
\section{Acknowledgments}
\vspace{-4pt}
This research was supported by the National Key Research And Development Program of China (2018YFC0830200,2018YFC0830206,
2019YFB1405802,2020YFC0832505).

\bibliographystyle{ACM-Reference-Format}
\balance
\bibliography{sample-base}
\end{document}